\def\year{2023}\relax
\documentclass[letterpaper]{article} 
\usepackage{array} 
\usepackage{aaai22}  
\usepackage{times}  
\usepackage{helvet}  
\usepackage{courier}  
\usepackage[hyphens]{url}  
\usepackage[colorlinks=true,urlcolor=blue,citecolor=black,linkcolor = black]{hyperref} 
\usepackage{graphicx} 
\usepackage{natbib}  
\usepackage{caption} 
\DeclareCaptionStyle{ruled}{labelfont=normalfont,labelsep=colon,strut=off} 
\usepackage{algorithm}
\usepackage{algorithmic}

\usepackage{xspace}
\usepackage{booktabs}
\usepackage{amsmath}
\usepackage{amssymb}

\usepackage{newfloat}
\usepackage{listings}
\floatstyle{ruled}
\newfloat{listing}{tb}{lst}{}
\floatname{listing}{Listing}
\title{chatClimate: Grounding Conversational AI in Climate Science}
\author{
    Saeid Ashraf Vaghefi,\textsuperscript{\rm 1,2}
    Qian Wang,\textsuperscript{\rm 1}
    Veruska Muccione,\textsuperscript{\rm 1}
    Jingwei Ni,\textsuperscript{\rm 3}
    Mathias Kraus,\textsuperscript{\rm 4}
    Julia Bingler,\textsuperscript{\rm 5,6}
    Tobias Schimanski,\textsuperscript{\rm 1}
    Chiara Colesanti-Senni,\textsuperscript{\rm 1,5}
    Dominik Stammbach,\textsuperscript{\rm 3}
    Nicolas Webersinke,\textsuperscript{\rm 4}
    Christrian Huggel,\textsuperscript{\rm 1}
    Markus Leippold\textsuperscript{\rm 1,7}
}
\affiliations{
    \textsuperscript{\rm 1}University of Zurich, Switzerland\\
    \textsuperscript{\rm 2}Eawag, Switzerland\\
    \textsuperscript{\rm 3}ETH Zurich, Switzerland\\
    \textsuperscript{\rm 4}FAU Erlangen-Nuremberg, Germany\\
    \textsuperscript{\rm 5}University of Oxford, United Kingdom\\
    \textsuperscript{\rm 6}Council on Economic Policies\\
    \textsuperscript{\rm 7}Swiss Finance Institute (SFI)\\    saeid.vaghefi@geo.uzh.ch, 
    qian.wang@uzh.ch, veruska.muccione@geo.uzh.ch, jingweini8@gmail.com, mathias.kraus@fau.de, julia.bingler@smithschool.ox.ac.uk, tobias.schimanski@bf.uzh.ch, chiara.colesantisenni@bf.uzh.ch, dominik.stammbach@gess.ethz.ch, nicolas.webersinke@fau.de, christian.huggel@geo.uzh.ch, markus.leippold@bf.uzh.ch

}

\begin{document}

\maketitle

\begin{abstract}
Large Language Models (LLMs) have made significant progress in recent years, achieving remarkable results in question-answering tasks (QA). However, they still face two major challenges: hallucination and outdated information after the training phase. These challenges take center stage in critical domains like climate change, where obtaining accurate and up-to-date information from reliable sources in a limited time is essential and difficult. To overcome these barriers, one potential solution is to provide LLMs with access to external, scientifically accurate, and robust sources (long-term memory) to continuously update their knowledge and prevent the propagation of inaccurate, incorrect, or outdated information. In this study, we enhanced GPT-4 by integrating the information from the Sixth Assessment Report of the Intergovernmental (IPCC AR6), the most comprehensive, up-to-date, and reliable source in this domain. We present our conversational AI prototype, available at \url{www.chatclimate.ai},\footnote{We would like to express our gratitude to Francesco Leonetti from \url{www.frigg.eco} for his invaluable and voluntary support in setting up the server. The server will become available by mid-April.} and demonstrate its ability to answer challenging questions accurately in three different QA scenarios: asking from 1) GPT-4, 2) chatClimate, and 3) hybrid chatClimate. The answers and their sources were evaluated by our team of IPCC authors, who used their expert knowledge to score the accuracy of the answers from 1 (very-low) to 5 (very-high). The evaluation showed that the hybrid chatClimate provided more accurate answers, highlighting the effectiveness of our solution. This approach can be easily scaled for chatbots in specific domains, enabling the delivery of reliable and accurate information.
\end{abstract}

\section{Introduction}

\begin{figure*}[ht]
	\centering
	\includegraphics[width=0.8\textwidth]{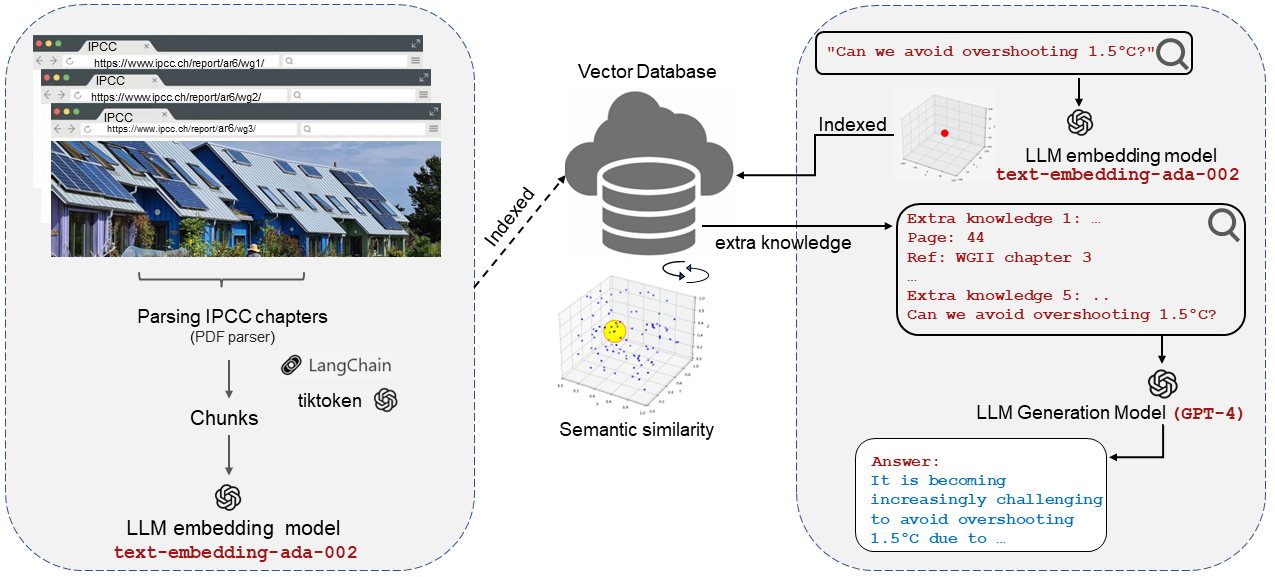}
	\caption{chatClimate Data Pipeline: from creating external memory, receiving questions to accurate answers from IPCC AR6.}
	\label{fig1:chatIPCC}
\end{figure*}

\textbf{Motivation.} 
Large Language Models (LLMs) have emerged as a promising area of research in Natural Language Processing (NLP) in recent years and have revolutionized text processing across various tasks, bringing significant advancements in natural language understanding and generation \cite{devlin2019bert,radford2019language,ouyang2022training}. Models such as LLaMA \cite{touvron2023llama}, T0 \cite{sanh2021multitask}, PaLM \cite{chowdhery2022palm}, GPT-3 \cite{brown2020language}, GPT-4 \cite{gpt4techreport}, or instruction fine-tuned models, such as ChatGPT \cite{instructgpt} and HuggingGPT \cite{shen2023hugginggpt}, have demonstrated their exceptional capabilities in generating human-like text across various domains, including language translation, summarization, and question answering.

LLMs can be used for Question Answering (QA) tasks, which require models to answer questions with no context given \cite{li2022selfprompting}. LLMs like GPT-3/3.5 have achieved impressive results on multiple choice question answering (MCQA) tasks in the zero, one, and few-shot settings\cite{robinson2023leveraging}. Recent works have used LLMs such as GPT-3 \cite{brown2020language} as an implicit knowledge engine to acquire the necessary knowledge for answering questions\cite{shao2023prompting}.

However, despite their impressive performances, LLMs suffer from two major issues: hallucination and outdated information after training \cite{Ji_2023}. These issues could be particularly problematic in domains such as climate change, where it is critical to have accurate, reliable, and timely information on changes in the climate systems, current impacts, and projected risks of climate change and solution space. Hence, providing accurate and up-to-date responses with authoritative references and citations is paramount to understanding the scale and immediacy of the climate crisis to facilitate the implementation of appropriate mitigation strategies. Enhanced communication between government entities and the scientific community fosters more effective dialogue between national delegations and policymakers. A facilitated feedback loop is established by guaranteeing the accuracy of information sources and responses, promoting informed decision-making in relevant domains. For example, governments may ask for feedback on specific statements in the report or request literature to support a claim. The importance of accurate and up-to-date information has been highlighted in previous studies as well \cite{bingler2022cheap,kumar2021climate,sethi2020importance}.
 
 By overcoming outdated information and hallucination challenges, LLMs can be used to extract relevant information from large amounts of text and assist in decision-making. Since training these models is computationally expensive and has many downsides \cite{FLI2023}, one possible solution is to provide the LLMs with external sources of information (called Long-term memory) to continuously update their knowledge and prevent the propagation of incorrect or outdated information. Several studies have explored the use of external data sources to enhance the performance of NLP models in specific domains \cite{kraus2023enhancing}. 

\textbf{Contribution.} In this paper, we introduce our prototype, chatClimate  (\url{www.chatclimate.ai}), a conversational AI designed to improve the veracity and timeliness of LLMs in the domain of climate change by utilizing the Sixth Assessment Report of the Intergovernmental Panel on Climate Change (hereafter IPCC AR6) \cite{ipcc2021climateWG1,ipcc2022climateWG2,ipcc2022climateWG3}.\footnote{As a disclaimer, please note that chatClimate is not endorsed by the IPCC and there is a high chance that chatClimate is hallucinating, may occasionally produce harmful instructions or biased content and may occasionally generate incorrect information.} These reports offer the latest and most comprehensive evaluation of the climate system, climate change impacts, and solutions related to adaptation, mitigation, and climate-resilient development. We evaluate the LLMs' performance in delivering accurate answers and references within the climate change domain by posing 13 challenging questions to our conversational AI (hereafter chatbot) across three scenarios: GPT-4, chatClimate standalone, and a hybrid chatClimate.

\textbf{Findings.} Our approach highlights the potential of incorporating external data sources to improve LLM performance in specific domains, such as climate change. By integrating up-to-date climate change information from the IPCC AR6 into LLMs, we can develop more accurate and reliable models for answering climate change-related questions. Our approach can potentially supply decision-makers and the public with trustworthy information on climate change, ultimately facilitating better-informed decision-making. This methodology can easily be scaled for chatbots in specialized domains to deliver more dependable and precise information.

The remainder of this paper is structured as follows. Section \ref{sec:background} gives a brief overview of recent developments in the field of LLMs, NLP in climate change, and the use of multiple sources in the context of LLMs. Section \ref{sec:Method} describes the method with which we develop our prototype, and Section \ref{sec:experiments} showcases example questions that are asked to access information from IPCC AR6. Section \ref{sec:discussion} discusses these results and the implications of this work. Section \ref{sec:conclusion} concludes this work.   

\section{Background}
\label{sec:background}
\subsection{Large Language Models}
LLMs have transformed NLP and AI research over the last few years \cite{fan2023bibliometric}. They show surprising new capabilities to generate creative text, solve basic math problems, answer reading comprehension questions, and more. These models primarily use transformer architecture and are trained on vast quantities of text data to identify patterns and connections within the data. Some notable examples of these models include GPT and BERT family models, which have been widely used for various NLP tasks \cite{devlin2019bert,radford2019language, brown2020language, ouyang2022training, gpt4techreport}. The recent breakthroughs with models like T0 \cite{sanh2021multitask}, LLaMA \cite{touvron2023llama}, PaLM \cite{chowdhery2022palm}, GPT-3 \cite{brown2020language}, and GPT-4 \cite{gpt4techreport} have further highlighted the potential of LLMs, with applications including chatbots \cite{, instructgpt} and virtual assistants \cite{jo2023promise}. However, LLMs can suffer from hallucination, which refers to mistakes in the generated text that are semantically incorrect or unsupported by the input text. This can lead to vague or inaccurate responses to questions. Moreover, most of these models are trained until the end of 2021 and have not been updated with new data or information since then, which can lead to outdated information \cite{Ji_2023}.

\subsection{NLP and Climate Change}
NLP techniques have been widely used in the analysis of text related to climate change. Applications range from financial climate disclosure analyses \cite{bingler2022cheap, luccioni2020analyzing}, topic modeling at the intersection of climate and health \cite{callaghan2021machine}, to climate claims fact-checking \cite{diggelmann2020climate, webersinke2022climatebert}.

\subsection{Question Answering and chatBots}
Question-answering (QA) systems and chatbots have become increasingly popular in various domains as they can provide users with relevant and accurate information conversationally. The importance, limitations, and future perspectives of conversational AI have been addressed in the literature from the open domain \cite{openai2022chatgpt, adiwardana2020humanlike} to domain-specific chatbots \cite{lin2023healthy}. When presented with a question in human language, chatbots can automatically provide a response. Although numerous information retrieval chatbots accomplish this task, deep learning has recently gained significant interest for its ability to learn the optimal representation tailored to QA tasks enabling the development of more sophisticated and context-aware QA systems and chatbots \cite{radford2019language, brown2020language, gpt4techreport, instructgpt}. In the context of climate change, QA systems and chatbots can help bridge the gap between complex scientific information and public understanding by providing concise and accessible answers to climate-related questions. Such systems can also facilitate communication between experts, policymakers, and stakeholders, enabling more informed decision-making and promoting climate change mitigation and adaptation strategies \cite{stammbach2022dataset,callaghan2021machine}. As the field of NLP and its application to climate change \cite{kolbel2020ask, bingler2022cheap} continues to advance, it is expected that QA systems and chatbots will play an increasingly important role in disseminating climate change information and fostering public engagement with climate science.
\subsection{Long-term Memory and Agents for LLMs}
One solution for enhancing the capabilities of LLMs in QA tasks is to fine-tune them on different datasets, which could be resource-wise expensive \cite{adiwardana2020humanlike}. However, an alternative approach involves using agents that access the LLMs' long-term memory, retrieve information, and insert it into a prompt to guide the LLMs more effectively \cite{kraus2023enhancing, nair2023dera}. These agents can decide which actions to perform, such as utilizing various tools, observing their outputs, or providing responses to user queries \cite{schick2023toolformer}. This approach has been shown to improve the accuracy and efficiency of LLMs in a range of domains, including healthcare and finance \cite{schick2023toolformer}. Domain-specific chatbots also use a similar concept, where an agent accesses an in-house database (long.term memory) to answer domain-specific questions \cite{gerhardyoung2022lowresource}. These chatbots can provide customized responses based on the available information in their database, allowing for more accurate and relevant answers to user queries.

\section{Method}
\label{sec:Method}

\subsection{chatClimate Pipeline}
In this study, we develop a long-term memory database by transforming the IPCC AR6 reports into a searchable format. The reports are converted from PDFs to JSON format, and each record in the database is divided into smaller chunks that LLM can easily process. The choice of the batch size for embeddings is a hyperparameter that requires tuning. We utilize OpenAI's state-of-the-art text embedding model to vectorize each data chunk. Prior to injection into the database, we implement an efficient indexing mechanism to optimize retrieval times and facilitate effective information retrieval. Consequently, we can implement a semantic search that identifies the most relevant results based on the meaning and context of each query.

When a user poses a question, it is first embedded and then indexed using semantic similarity to find the five nearest vectors corresponding to the inquiry. The dot product of two vectors is utilized to analyze the similarity between vector embeddings, which is obtained by multiplying their respective components and summing the results.

After identifying the nearest vectors to the query vector, they are decoded into text format. This information is then used to refine and improve prompts. Augmented queries are posed to the GPT-4 model through instructed prompts, which enhance the user experience and increase the overall performance of our chatBot. Figure \ref{fig1:chatIPCC} shows the pipeline of chatClimate. 

\subsection{Tools and External APIs}
The first tool used in this study is a Python-based module that transforms IPCC AR6 reports from PDFs to JSON format (PDF parsr) and preprocesses the data, utilizing the powerful pandas library to access and manipulate data stored in dataframes.

The second tool is the LangChain Python package (https://github.com/hwchase17/langchain), which retrieves data from the JSON and chunks the extracted text into smaller sizes, ready for embedding. LangChain is a lightweight layer that transforms sequential LLM interactions into a natural conversation experience.

The third tool employed is OpenAI's embedding model "text-embedding-ada-002," which vectorizes all chunks of the IPCC AR6 JSON files. Vector embeddings have proven to be a valuable tool for a variety of machine learning applications, as they can efficiently represent objects as dense vectors containing semantic information. 

The fourth tool involves storing the generated vectors in a database, allowing for efficient storage and retrieval of the vector embeddings.

The fifth and final tool used is the GPT-4 "chatcompletion" endpoint with instructed prompts, which provides answers to questions by leveraging the indexed vector embeddings.

\subsection{Input Prompts and chatBots}
The importance of prompt engineering for LLMs has been addressed in previous work \cite{kraus2023enhancing}. We designed three prompts to compare the answers of our chatbots (i.e., chatClimate with  hybrid chatClimate and GPT-4). The prompt used in our study consists of a series of instructions that guide the completion of a chat with GPT-4 on how to answer a provided question. The prompt is structured to allow the chatbot to access external resources while using its in-house knowledge.

In the first scenario, hybrid chatClimate: The prompt starts with five pieces of external information retrieved from long-term memory, followed by a question that was asked by the user. The prompt instructs the chatbot to provide an answer based on the given information while using its own knowledge. Moreover, the chatbot is structured to prioritize IPCC AR6 for answers, referencing the names and pages of corresponding IPCC reports (Working Group I, II, III chapters, summary for policymakers, technical summary, and synthesis reports).
In the second scenario, chatClimate: The prompt starts with five pieces of external information retrieved from long-term memory, followed by a question that was asked by the user. The prompt instructs the chatbot to provide answers only based on IPCC AR6.
In the third scenario, GPT-4: The prompt does not provide any extra information or instruction on how to provide answers.

Overall, the prompt is designed to guide how to answer the questions given the availability of external and/or in-house knowledge.
In the following, we demonstrate the three prompts used in this study:
Listing\ref{code:prompt_1}, \ref{code:prompt_2} and \ref{code:prompt_3}.

\begin{lstlisting}[frame=single, basicstyle=\ttfamily\footnotesize, xleftmargin=0pt, numbers=none, label=code:prompt_1, caption={Input prompt for hybrid chatClimate.}]
"system", "content":  You are a Q&A bot, an intelligent system that answers user questions based on the information provided by the user above the question and your in-house knowledge. There are 5 pieces of extra information above the user question. please indicate the Page and Reference, which are provided below each piece of information. Additionally, let us know which part of your answer is from the IPCC information and which part is based on your in-house knowledge by writing either (IPCC AR6) or (In-house knowledge). If the information cannot be found in the information provided by the user or your in-house knowledge, please say 'I don't know'.
"role": "user", "content": External knowledge + Question 
\end{lstlisting}

\begin{lstlisting}[frame=single, language=Python, basicstyle=\ttfamily\footnotesize, xleftmargin=0pt, numbers=none, label=code:prompt_2, caption={Input prompt for chatClimate.}]
"system", "content":  You are a Q&A bot, an intelligent system that answers user questions ONLY based on the information provided by the user. If you use user information, please indicate the Page and Reference, which are provided below each piece of information. If the information cannot be found in the information provided by the user, please say 'I don't know'
"role": "user", "content": External knowledge + Question 
\end{lstlisting}

\begin{lstlisting}[frame=single, language=Python, basicstyle=\ttfamily\footnotesize, xleftmargin=0pt, numbers=none, label=code:prompt_3, caption={Input prompt for GPT-4}]
"system", "content": "You are Q&A bot. A highly intelligent system that answers user questions" 
"role": "user", "content": Question 
\end{lstlisting}

\section{Experiments}
\label{sec:experiments}
\subsection{chatBots and Questions}
We conducted three sets of experiments by asking hybrid chatClimate, chatClimate, and GPT-4 chatbots 13 questions \ref{tbl:table_1}. Our team of IPCC AR6 authors then assessed the answers' accuracy. It's worth noting that our prototype's ability to provide sources for statements can facilitate the important process of trickle-back, which is often required by governments and other stakeholders in the context of IPCC reports.

\begin{table}[hbt!]
\caption{The 13 designed questions for running the 3 sets of experiments}
\label{tbl:table_1}
\centering
\scriptsize
\renewcommand\arraystretch{1}
\newcolumntype{C}{>{\centering\arraybackslash}p{3cm}} 
\begin{tabular}{p{4.3cm}C}
\toprule
\textbf{Questions} & \textbf{Difficulty (1-5)} \\
\textbf{} & 1:very-low, 2:low, 3:medium, 4:high, 5:very-high \\
\midrule
Q1= "Is it still possible to limit warming to 1.5°C?" & 3\\
Q2= "When will we reach 1.5°C?" & 3\\
\midrule
Q3= "What does overshoot mean?" & 1\\
Q4= "Can we avoid overshooting 1.5°C?" & 3\\
\midrule
Q5= "Have emissions reductions fallen for some countries?" & 3\\
\midrule
Q6= "What are the issues with financing adaptation?" & 4\\
Q7= "Where is the majority of climate finance going?" & 4\\
\midrule
Q8= "What are the options for scaling up adaptation and mitigation in developing countries?" & 4\\
\midrule
Q9= "Which regions will be disproportionally affected by climate change?" & 5\\
\midrule
Q10= "What is climate justice?" & 2\\
\midrule
Q11= "What is maladaptation?" & 4\\
Q12= "Is there evidence of maladaptation?" & 5\\
Q13= "Will glaciers in Scotland melt?" & 5\\
\bottomrule
\end{tabular}
\end{table}

\begin{table*}[hbt!] 
\caption{Comparison of generated answers to questions 1 and 2 from hybrid chatClimate, chatClimate, and GPT-4}
\label{tbl:table_2}
\centering
\scriptsize
\begin{tabular}{p{5.3cm}p{5.3cm}p{5.3cm}}
\toprule
\multicolumn{3}{c}{\textbf{QA comparision}} \\
\textbf{hybrid chatClimate} & \textbf{chatClimate}  & \textbf{GPT-4} \\
\midrule
\multicolumn{3}{c}{\textbf{Q1) Is it still possible to limit warming to 1.5°C?}}\\\\

Yes, it is still possible to limit global warming to 1.5°C, but it would require immediate, deep, rapid, and sustained reductions in greenhouse gas emissions, including CO2 and methane (Page:31, IPCC AR6 WGIII Chapter03). Scenarios limiting warming to 1.5°C with no or limited overshoot show that net energy system CO2 emissions must fall by 35-51\% by 2030 and 87-97\% by 2050 (Page:43, IPCC AR6 WGIII Technical Summary). Net electricity sector CO2 emissions would need to reach zero globally between 2045 and 2055 (Page:3, IPCC AR6 WGIII Chapter06). However, the likelihood of limiting warming to 1.5°C with no or limited overshoot has decreased compared to the Special Report on Global Warming of 1.5°C due to the rise in global greenhouse gas emissions since its publication (Page:4, IPCC AR6 WGIII Chapter03). Only a small number of published pathways achieve this target without overshooting over the 21st century (Page:31, IPCC AR6 WGIII Technical Summary). In summary, while still possible, limiting warming to 1.5°C would require immediate and drastic action to reduce greenhouse gas emissions (In-house knowledge). 
& It is still possible to limit warming to 1.5°C with no or limited overshoot, but the likelihood has decreased since the Special Report on Global Warming of 1.5°C (SR1.5) due to the rise in global GHG emissions since it was published. This has led to higher near-term emissions (2030) and higher cumulative CO2 emissions until the time of net zero (medium confidence). Limiting warming to 1.5°C requires deep, rapid, and sustained reductions in greenhouse gas emissions, including methane, alongside rapid reductions of CO2 emissions to net zero {IPCC AR6 WGIII Chapter03, Page:4, IPCC AR6 WGIII Technical Summary, Page:31}.
& Limiting global warming to 1.5°C is becoming increasingly challenging due to the continued rise in greenhouse gas emissions and the slow pace of implementing effective climate policies. However, it is not entirely impossible. Achieving this goal would require rapid and far-reaching transitions in energy, land, urban infrastructure, and industrial systems. This includes drastically reducing greenhouse gas emissions, increasing renewable energy sources, improving energy efficiency, and adopting sustainable land use practices. The Intergovernmental Panel on Climate Change (IPCC) has outlined various pathways to achieve this target, but it requires immediate and ambitious action from governments, businesses, and individuals worldwide.\\\\
\textbf{accuracy: 5 } & \textbf{accuracy: 5 } & \textbf{accuracy: 3 } \\
\midrule
\multicolumn{3}{c}{\textbf{Q2) When will we reach 1.5°C?}}\\\\
We are likely to reach 1.5°C of global warming between 2030 and 2052 if the current rate of warming continues (high confidence) (IPCC AR6, Reference: IPCC AR6 WGI Chapter01, Page: 44). However, the exact timing may vary depending on the emissions reduction pathways and efforts to achieve net-zero CO2 emissions (In-house knowledge). 
& We are likely to reach 1.5°C of global warming between 2030 and 2052 if it continues to increase at the current rate (high confidence). Reference: IPCC AR6 WGI Chapter01, Page: 44.
& As an AI, I cannot predict the exact year when global temperatures will reach 1.5°C above pre-industrial levels. However, according to the Intergovernmental Panel on Climate Change (IPCC), we could reach this threshold between 2030 and 2052 if global warming continues at its current rate. The actual timeline depends on various factors, including greenhouse gas emissions, climate policies, and global efforts to mitigate climate change.\\\\
\textbf{accuracy: 5 } & \textbf{accuracy: 4 } & \textbf{accuracy: 2 } \\
\bottomrule
\end{tabular}
\end{table*}

Table \ref{tbl:table_2} presents the returned answers from chatbots. The question "Is it still possible to limit warming to 1.5°C?" targets the mitigation, and the hybrid chatBot and chatClimate explicitly return the greenhouse gas emission reduction amounts and time horizon while the GPT-4 answer is more general.

\subsection{Evaluation of Answers (accuracy score)}

Several studies focus on human-chatbot interaction effectiveness \cite{abdar2020impact, ramachandran2020framework, luger2016towards, PRZEGALINSKA2019785}. Evaluation involves factors such as relevance, clarity, tone, style, speed, consistency, personalization, error handling, and user satisfaction. This work, however, only examines the chatbot's performance on accuracy.

\begin{figure*}[hbt!]
	\centering
	\includegraphics[width=0.9\textwidth]{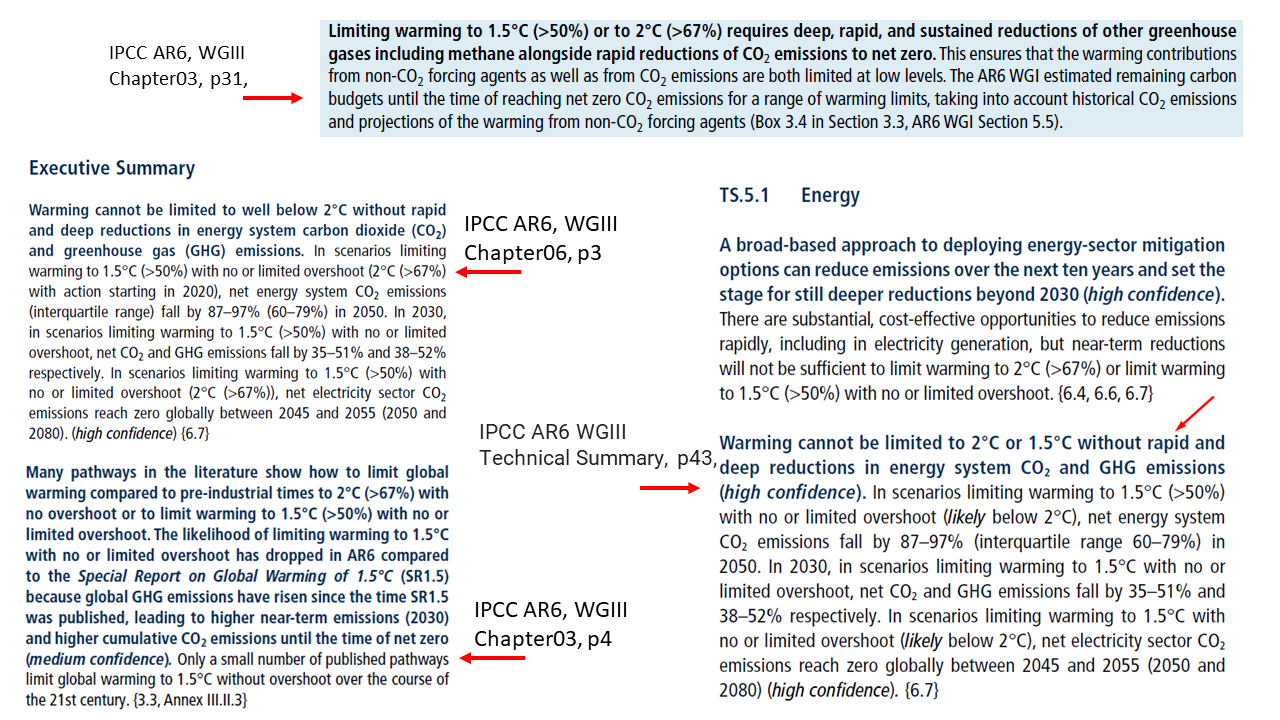}
	\caption{Cross-checking of the references for Question1}
	\label{fig2:crossCheckQ1}
\end{figure*}

\begin{figure*}[hbt!]
	\centering
	\includegraphics[width=0.9\textwidth]{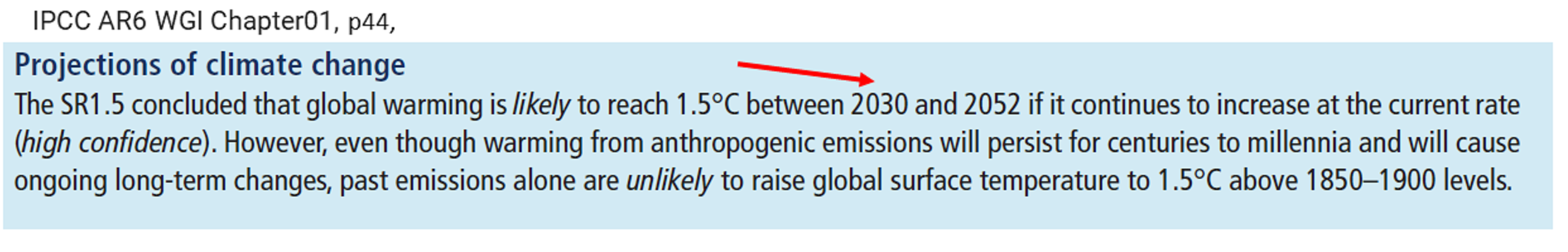}
	\caption{Cross-checking of the references for Question2}
	\label{fig3:crossCheckQ2}
\end{figure*}

\section{Discussion}
\label{sec:discussion}
\subsection{Expert Cross-Check of the Answers} 
Overall, the responses provided by the hybrid chatClimate were more accurate than those of chatClimate standalone and GPT-4. For the sake of brevity, we have provided a detailed analysis of Q1 and Q2 in Table \ref{tbl:table_2}, and only highlight the key issues for Q3-Q13. For instance, in Q1, we asked the bots whether it is still possible to limit global warming to 1.5°C. Both the hybrid chatClimate and chatClimate systems referred to the amount of CO2 that needs to be reduced over different time horizons to stay below 1.5°C. However, the GPT-4 provided a more general response. 
To verify the accuracy of responses generated by the chatClimate bots, we cross-checked the references provided by both systems. We found that the chatClimate bots consistently provided sources for their statements, as shown in Figure 2 and 3, which is essential for verifying the veracity of the bot's answer.
In Q2, we asked the bots about the time horizon when we reach 1.5°C. All three bots similarly referred to the time horizon of 2030 to 2052 based on the mitigation measures we take into account. The consistency of the answers shows that this time horizon has been mentioned in the training data of GPT-4 as well (e.g. IPCC AR6 WGI, which was released in August 2021 or Special Report of IPCC on Global Warming of 1.5 ºC, 2018).

\begin{table*}[hbt!]
\caption{Cross-checking of the hybrid chatClimate references for Question 9 with k-5, k-10, and k-15 nearest for extra knowledge retrieval}
\label{tbl:table_3}
\centering
\scriptsize
\begin{tabular}{p{5.3cm}p{5.3cm}p{5.3cm}}
\toprule
\multicolumn{3}{c}{\textbf{QA comparison}} \\
\textbf{Hybrid K-5} & \textbf{Hybrid K-10}  & \textbf{Hybrid K-15} \\
\midrule
\multicolumn{3}{c}{\textbf{Q6) Where is the majority of climate finance going?}}\\\\ 

The majority of climate finance is going to a few large economies, with Brazil, India, China, and South Africa accounting for around one-quarter to more than a third depending on the year, a share similar to that represented by developed countries (IPCC AR6, WGIII Chapter 15, Page 18). Additionally, global climate finance is heavily focused on mitigation, with more than 90\% on average between 2017-2020 being directed towards mitigation efforts (IPCC AR6, WGIII Chapter 15, Page 3).
&
Regions that will be disproportionally affected by climate change include East, Central, and West Africa, South Asia, Micronesia and Melanesia, and Central America (IPCC AR6, Page 81). These regions are highly vulnerable to climate hazards such as droughts, floods, and sea-level rise. In Asia, climate change hotspots with strong climate signals and high concentrations of vulnerable people are present in semiarid, glacial-fed river basins and mega deltas (IPCC AR6, Page 10). South Asia and Africa are also identified as hotspots of multi-sectoral climate change risk, largely due to changes in water-related indicators affecting crop production (IPCC AR6, Page 81).
Disproportionate impacts are exacerbated by factors such as poverty, hunger, conflict, and environmental degradation, which make countries and regions more vulnerable to climate change (IPCC AR6, Page 81). Vulnerable populations within these regions, including the poor, children, women, and marginalized Indigenous or ethnic minority people, often have limited resources and structures to support them during climate-related disasters (IPCC AR6, Page 27).
Regions such as Africa and Central and South America are more likely to experience disproportionately higher impacts due to the increasing frequency of extreme temperatures, which can affect public health, agriculture, ecosystems, biodiversity, and infrastructure (IPCC AR6, Page 68). In addition, these regions are projected to face increased risks of flood, hunger, and water scarcity (IPCC AR6, Pages 75, 81, 110).
In summary, regions that will be disproportionally affected by climate change include parts of Africa, South Asia, Central America, and specific climate-change hotspots in Asia. These regions face higher risks due to a combination of climate hazards, vulnerable populations, and limited resources for adaptation and support (In-house knowledge).
&
Regions that will be disproportionally affected by climate change include East, Central, and West Africa, South Asia, Micronesia and Melanesia, and Central America (IPCC AR6, WGII Chapter08, Page:81). These regions are highly vulnerable to climate hazards such as droughts, floods, and sea-level rise. In addition, Africa and Central and South America are more likely to experience disproportionately higher impacts due to their vulnerability, adaptation gaps, and development needs (IPCC AR6, WGII Chapter18, Page:68).
South Asia and Africa are also identified as hotspots of multi-sectoral climate change risk, largely due to changes in water-related indicators affecting crop production (IPCC AR6, WGII Chapter16, Page:81). Climate change is expected to increase the magnitude and frequency of droughts and hurricanes in these regions, potentially forcing millions of people to leave their homes (IPCC AR6, WGII Chapter12, Page:80).
Furthermore, vulnerable groups within these regions, such as the poor, children, women, and marginalized Indigenous or ethnic minority people, are differentially impacted by climate change (IPCC AR6, WGII Chapter08, Page:27). These vulnerable regions and countries often have limited resources and structures to support these groups when climate hazards occur, placing an additional burden on them (In-house knowledge).
In summary, regions such as East, Central, and West Africa, South Asia, Micronesia and Melanesia, and Central America will be disproportionally affected by climate change due to their high vulnerability, exposure to climate hazards, and limited resources to support vulnerable populations (IPCC AR6, WGII Chapter08, Page:81).\\
\textbf{accuracy: 3  } & \textbf{accuracy: 3 } & \textbf{accuracy: 4 } \\
\bottomrule
\end{tabular}
\end{table*}

\subsection{Prompts Engineering} By adjusting the retrieval hyperparameters, we can extract extra knowledge from long-term memory. We retrieved 10 and 15 nearest vectors to the user's query vector. With this, GPT-4 can obtain more information and provide more accurate answers when accessing the outside world. The results have been demonstrated in Table 3, where K-15 scenario has provided more information about the regions that are most affected by climate change. This, in particular, shows the importance of prompt engineering,

\subsection{Personalized-GPTs or GPT-n, Risk Management}
Domain-specific chatbots and conversational AI tools provide easy access to accurate information. However, potential risks from external data sources, such as inaccuracies or biases, should be acknowledged. In this study, we developed and implemented domain-specific chatbots for the climate change domain. We compared three chatbot scenarios and found that the hybrid chatClimate provided more accurate answers to 13 sample questions. We evaluated the answers internally, benefitting from the expert knowledge of co-authors. Since training LLMs is resource-intensive \cite{gpt4techreport}, integrating them with the outside world by providing long-term memory and prompt engineering could yield better results with fewer resources. However, creating long-term memory requires caution. We used the IPCC AR6 as a comprehensive and reliable source to build external memory for LLMs, highlighting the importance of such databases for chatbot accuracy. Although there is an ongoing debate about pausing LLM training for six months until proper regulations are established, we believe that regulating LLM training, fine-tuning, and incorporating it into applications is necessary. Specifically, external database integration and prompt engineering should be considered in regulations for chatbots.

\subsection{Limitations}
\textbf{chain of thoughts (COTs)}: In this study, we did not fully explore the potential of chain of thoughts (COTs) by testing different prompts. However, we expect that implementing COTs will improve the accuracy of our system's outputs which we plan in our future works.\\

\subsection{Future works} We continually improve chatClimate and welcome community feedback on our website \url{www.chatclimate.ai/ipcc} to enhance its question-answering capabilities. Our goal is to provide accurate and reliable climate change information, and we believe domain-specific chatbots like chatClimate play a crucial role in achieving it.

\section{Conclusion}
\label{sec:conclusion}
In summary, our study demonstrates that the hybrid chatClimate outperformed both GPT-4 and chatClimate standalone in terms of the accuracy of answers when provided access to the outside world (IPCC AR6). The higher performance can be attributed to the integration of up-to-date and domain-specific data, which addresses the issues of hallucination and outdated information often encountered in LLMs. The results underline the importance of tailoring models to specific domains. The main findings of our work are summarized as follows:
\begin{enumerate}
    \item The hallucination and outdated issues of LLMs could be refined by giving those models access to the knowledge beyond their training phase time and instructing LLMs on how to utilize that knowledge.
    \item With prompt engineering and knowledge retrieval, LLMs can provide sources of the answers properly.
    \item Hyperparameter tuning during knowledge retrieval and semantic search plays an important role in prompt engineering. We tested this by K-5, K-10, and K-15 nearest pieces of knowledge to the question in the semantic search between the question and the database.
    \item Regulating LLM training, fine-tuning, and incorporating it into applications are necessary. Specifically, external database integration and prompt engineering should be considered in regulations for chatbots. We emphasize the importance of regulation for checking the outcomes of domain-specific chatbots. In such domains, users may not have enough knowledge to verify answers or cross-check references, making biased data or engineered prompts potentially harmful to end users.
    \item Our findings not only emphasize the importance of leveraging climate domain information in QA tasks but also highlight the need for continued research and development in the field of AI-driven text processing.
\end{enumerate}

\bibliography{acl2023}



\clearpage
\onecolumn
\section*{Appendix}
\label{sec:Appendix}
\subsection{Expert cross-check of the QA scenarios Q3-13}
Here we continue the demonstration of QA and our analyses for questions 3 to 13.\\

\begin{table*}[hbt!]
\label{tbl:SM1}
\caption*{Table SM1: Comparison of generated answers from hybrid chatClimate, chatClimate, and GPT-4 to question3}
\centering
\scriptsize
\begin{tabular}{p{5.3cm}p{5.3cm}p{5.3cm}}
\toprule
\textbf{hybrid chatClimate} & \textbf{chatClimate}  & \textbf{GPT-4} \\
\midrule
\multicolumn{3}{c}{\textbf{Q3) What does overshoot mean?}}\\\\
Overshoot refers to the exceedance of a specified global warming level, followed by a decline to or below that level during a specified period of time (e.g., before 2100). In the context of climate change, overshoot pathways are scenarios where a certain concentration, forcing, or global warming level is first exceeded and then returns to or below that level before the end of a specified period of time. The overshoot duration can vary, but in most cases, it occurs over a period of at least one decade and up to several decades (IPCC AR6 WGI AnnexVII, Page:28; IPCC AR6 WGIII Annex I, Page:20). 

& Overshoot refers to the exceedance of a specified global warming level, followed by a decline to or below that level during a specified period of time (e.g., before 2100). Sometimes the magnitude and likelihood of the overshoot are also characterized. The overshoot duration can vary from one pathway to the next, but in most overshoot pathways in the literature and referred to as overshoot pathways in the AR6, the overshoot occurs over a period of at least one decade and up to several decades. See also Pathways. Reference: IPCC AR6 WGII Annex-II, Page: 29. 

& Overshoot refers to a situation where a value or process exceeds its intended target or goal, often temporarily, before stabilizing or returning to the desired level. In various contexts, it can refer to:
\begin{enumerate}
    \item In ecology, overshoot occurs when a population exceeds the carrying capacity of its environment, leading to a decline in resources and potential population collapse.
    \item In economics, overshoot can refer to a situation where economic growth or inflation exceeds its sustainable level, leading to a subsequent correction or downturn.
    \item In engineering and control systems, overshoot refers to the transient response of a system that exceeds its final steady-state value before settling down to that value.
    \item In physics, overshoot can refer to a situation where an object moves beyond its intended stopping point, often due to momentum or external forces, before returning to the desired position.
    
\end{enumerate}
\\
\textbf{accuracy: 5 } & \textbf{accuracy: 4 } & \textbf{accuracy: 2 } \\
\bottomrule
\end{tabular}
\end{table*}
\clearpage
\begin{table*}[hbt!]
\caption*{Table SM2: Comparison of generated answers from hybrid chatClimate, chatClimate, and GPT-4 to question4}
\label{tbl:SM2}
\centering
\scriptsize
\begin{tabular}{p{5.3cm}p{5.3cm}p{5.3cm}}
\toprule
\textbf{hybrid chatClimate} & \textbf{chatClimate}  & \textbf{GPT-4} \\
\midrule
\multicolumn{3}{c}{\textbf{Q4) Can we avoid overshooting 1.5°C?}}\\\\
Avoiding overshooting 1.5°C is challenging, but not impossible. The likelihood of limiting warming to 1.5°C with no or limited overshoot has dropped in AR6 compared to the Special Report on Global Warming of 1.5°C (SR1.5) because global GHG emissions have risen since the time SR1.5 was published, leading to higher near-term emissions (2030) and higher cumulative CO2 emissions until the time of net zero (medium confidence) (IPCC AR6 WGIII Chapter03, Page:4). Only a small number of published pathways limit global warming to 1.5°C without overshoot over the course of the 21st century (IPCC AR6 WGIII Chapter03, Page:4).

Mitigation pathways with 2030 emissions levels in the NDCs announced prior to COP26 consistently overshoot 1.5°C by 0.15°C-0.3°C (IPCC AR6 WGIII Chapter03, Page:60). To avoid overshooting 1.5°C, it would require rapid and significant reductions in greenhouse gas emissions, as well as the implementation of effective climate policies and technologies (In-house knowledge). 

&Based on the information provided, avoiding overshooting 1.5°C is challenging. The likelihood of limiting warming to 1.5°C with no or limited overshoot has dropped in AR6 compared to the Special Report on Global Warming of 1.5°C (SR1.5) because global GHG emissions have risen since the time SR1.5 was published, leading to higher near-term emissions (2030) and higher cumulative CO2 emissions until the time of net zero (medium confidence). Only a small number of published pathways limit global warming to 1.5°C without overshoot over the course of the 21st century (IPCC AR6 WGIII Chapter03, Page:4).

& As an AI, I cannot predict the future, but I can provide information on the current situation. Avoiding overshooting 1.5°C of global warming above pre-industrial levels is a challenging task, as we have already experienced around 1.1°C of warming. To avoid overshooting 1.5°C, rapid and far-reaching actions in energy, land use, urban planning, transportation, and industrial systems are required. This would involve transitioning to renewable energy, increasing energy efficiency, reducing deforestation, and adopting sustainable agriculture practices, among other measures. The Intergovernmental Panel on Climate Change (IPCC) has outlined various pathways to achieve this goal, but it requires immediate and ambitious efforts from governments, businesses, and individuals worldwide.\\
\textbf{accuracy: 5 } & \textbf{accuracy: 4 } & \textbf{accuracy: 3 } \\

\bottomrule
\end{tabular}
\end{table*}
\clearpage
\begin{table*}[hbt!]
\caption*{Table SM3: Comparison of generated answers from hybrid chatClimate, chatClimate, and GPT-4 to question5}
\label{tbl:SM_Table3_Q5}
\centering
\scriptsize
\begin{tabular}{p{5.3cm}p{5.3cm}p{5.3cm}}
\toprule
\textbf{hybrid chatClimate} & \textbf{chatClimate}  & \textbf{GPT-4} \\
\midrule
\multicolumn{3}{c}{\textbf{Q5) Have emissions reductions fallen for some countries?}}\\\\
Yes, emissions reductions have fallen for some countries. A growing number of countries have reduced CO2 and GHG emissions for longer than 10 years (high confidence). Data up to 2018 indicates that about 24 countries have reduced territorial CO2 and GHG emissions (excluding LULUCF CO2), as well as consumption-based CO2 emissions, for at least 10 years (Page 22, IPCC AR6 WGIII Chapter02). At least 18 countries have sustained production-based GHG and consumption-based CO2 emission reductions for longer than 10 years (Page 11, IPCC AR6 WGIII SummaryForPolicymakers; Page 13, IPCC AR6 WGIII SPM). These reductions were linked to energy supply decarbonisation, energy efficiency gains, and energy demand reduction, which resulted from both policies and changes in economic structure (Page 11, IPCC AR6 WGIII SummaryForPolicymakers; Page 13, IPCC AR6 WGII SPM). However, the total reduction in annual GHG emissions of these countries is small (about 3.2 GtCO2-eq per yr) compared to global emissions growth observed over the last decade (Page 3, IPCC AR6 WGIII Chapter02).
& Yes, a growing number of countries have achieved GHG emission reductions for longer than 10 years. Data up to 2018 indicates that about 24 countries have reduced territorial CO2 and GHG emissions (excluding LULUCF CO2), as well as consumption-based CO2 emissions, for at least 10 years. At least 18 countries have sustained production-based GHG and consumption-based CO2 emission reductions for longer than 10 years. These reductions were linked to energy supply decarbonization, energy efficiency gains, and energy demand reduction, which resulted from both policies and changes in economic structure. However, these reductions have only partly offset global emissions growth (Reference: IPCC AR6 WGIII Chapter02, Page:22; IPCC AR6 WGIII SummaryForPolicymakers, Page:11; IPCC AR6 WGIII SPM, Page:13; IPCC AR6 WGIII Chapter02, Page:3).
& Yes, emissions reductions have fallen for some countries over the years. Various factors contribute to this, such as the implementation of climate policies, increased use of renewable energy sources, improvements in energy efficiency, and economic shifts. The extent of emissions reductions varies across countries, depending on their specific circumstances and efforts to combat climate change. However, it is important to note that global greenhouse gas emissions continue to rise, and more collective action is needed to achieve significant reductions on a global scale.\\
\textbf{accuracy: 5} & \textbf{accuracy: 4} & \textbf{accuracy: 3 } \\
\bottomrule
\end{tabular}
\end{table*}
\clearpage
\begin{table*}[hbt!]
\caption*{Table SM4: Comparison of generated answers from hybrid chatClimate, chatClimate, and GPT-4 to question6}
\label{tbl:SM_Table4_Q6}
\centering
\scriptsize
\begin{tabular}{p{5.3cm}p{5.3cm}p{5.3cm}}
\toprule
\textbf{hybrid chatClimate} & \textbf{chatClimate}  & \textbf{GPT-4} \\
\midrule
\multicolumn{3}{c}{\textbf{Q6) What are the issues with financing adaptation?}}\\\\
There are several issues with financing adaptation, including:
\begin{enumerate}
    \item Implicit economic world views and technical support availability: The choice of funding and financing mechanisms is often based on implicit economic world views or the technical support available to sub-national governments, such as preparing municipal bonds or contracting for public-private partnerships (IPCC AR6, WGII Chapter 6, Page 70).
    \item Justice implications: Adaptation finance has profound justice implications, as it can intensify financial and environmental risks borne primarily by the poor, the working class, or people discriminated against because of race, sexual orientation, or ability (IPCC AR6, WGII Chapter 6, Page 70).
    \item Dwindling resources and trade-offs: Discussions of adaptation finance often occur in the context of dwindling resources and trade-offs, leading to triage decisions about other investments that societies can tolerate suspending (IPCC AR6, WGII Chapter 8, Page 78).
    \item Governance challenges: In many poor, vulnerable countries, complex governance challenges, such as budget austerity or corruption, hamper the provision of support for adaptation (IPCC AR6, Chapter 8, Page 78).
    \item Insufficient financing: Insufficient financing is a key driver of adaptation gaps, with annual finance flows targeting adaptation being significantly less than the lowest adaptation cost estimates for near-term climate change (IPCC AR6, WGII Technical Summary, Page 50).
    \item Limited private-sector financing: Globally, private-sector financing of adaptation has been limited, especially in developing countries (IPCC AR6 WGII, Technical Summary, Page 50).
    \item Weak incentives for climate-resilient financial strategies: The use of climate-resilient financial strategies and instruments is limited by weak incentives, high upfront costs, high transaction and intermediation costs, and relatively long pay-off times (IPCC AR6, WGII Chapter 5, Page 113).
    \item Imbalanced allocation of public and private finance: Given limited experience and weak information for adaptation finance, sub-optimal outcomes may include imbalanced allocation of public and private finance, as well as inequitable division of risks and returns (IPCC AR6, WGII Chapter 5, Page 113).
\end{enumerate}
& There are several issues with financing adaptation:
\begin{enumerate}
    \item Implicit economic world views and technical support availability influencing the choice of funding and financing mechanisms (Page 70, IPCC AR6 WGII Chapter06).
    \item Adaptation finance having profound justice implications, as it can intensify financial and environmental risks primarily borne by the poor, working class, or people discriminated against due to race, sexual orientation, or ability (Page 70, IPCC AR6 WGII Chapter06).
    \item Dwindling resources and trade-offs, leading to triage decisions about other investments that societies can tolerate suspending (Page 78, IPCC AR6 WGII Chapter08).
    \item Complex governance challenges, such as budget austerity or corruption, hampering the provision of support in many poor, vulnerable countries (Page 78, IPCC AR6 WGII Chapter08).
    \item In the absence of adaptation funding for the poor, the costs of adaptation are borne by the poor at community, kin-group, and household scales, which can ultimately increase the likelihood of remaining trapped in poverty (Page 78, IPCC AR6 WGII Chapter08).
    \item Insufficient financing being a key driver of adaptation gaps, with finance not targeting more vulnerable countries and communities (Page 50, IPCC AR6 WGII TechnicalSummary)
    \item Limited use of climate-resilient financial strategies and instruments due to weak incentives, high upfront costs, high transaction and intermediation costs, and relatively long pay-off times (Page 113, IPCC AR6 WGII Chapter05).
    \item Imbalanced allocation of public and private finance, as well as inequitable division of risks and returns (Page 113, IPCC AR6 WGII Chapter05).
\end{enumerate} 
& Financing adaptation to climate change involves several challenges and issues, including: 
\begin{enumerate} 
    \item Insufficient funding: The amount of funding available for adaptation projects is often inadequate to meet the growing needs of countries and communities vulnerable to climate change impacts.
    \item Uncertainty in climate change impacts: The unpredictable nature of climate change makes it difficult to accurately estimate the costs of adaptation measures, which can lead to underfunding or misallocation of resources.
    \item Prioritization of adaptation projects: With limited funding, it is challenging to prioritize adaptation projects based on their urgency, effectiveness, and potential benefits, leading to potential inefficiencies in resource allocation.
    \item Lack of access to finance: Developing countries, particularly the least developed and most vulnerable, often face difficulties in accessing international climate finance due to complex procedures, lack of capacity, and limited resources.
    \item Inadequate integration of adaptation into development planning: Adaptation measures are often not well-integrated into national and local development plans, leading to a lack of coherence and coordination in addressing climate change impacts.
    \item Limited private sector involvement: The private sector has a crucial role to play in financing adaptation, but barriers such as perceived risks, lack of incentives, and limited awareness of opportunities often hinder their engagement.
    \item Short-term focus: Funding for adaptation is often focused on short-term projects, while long-term planning and investments are needed to build resilience to climate change.
    \item Monitoring and evaluation challenges: Assessing the effectiveness of adaptation projects is difficult due to the lack of standardized metrics, methodologies, and data, which can hinder learning and improvement in adaptation planning and financing.
    \item Capacity constraints: Many developing countries lack the technical, institutional, and financial capacity to plan, implement, and monitor adaptation projects effectively.
    \item Equity concerns: Financing for adaptation should be distributed fairly, taking into account the needs and vulnerabilities of different countries and communities. However, this can be challenging due to competing priorities, limited resources, and political considerations.
\end{enumerate}.\\
\textbf{accuracy: 4} & \textbf{accuracy: 4 } & \textbf{accuracy: 3 } \\
\bottomrule
\end{tabular}
\end{table*}
\clearpage
\begin{table*}[hbt!]
\caption*{Table SM5: Comparison of generated answers from hybrid chatClimate, chatClimate, and GPT-4 to question7}
\label{tbl:SM_Table5_Q7}
\centering
\scriptsize
\begin{tabular}{p{5.3cm}p{5.3cm}p{5.3cm}}
\toprule
\multicolumn{3}{c}{\textbf{QA comparison}} \\
\textbf{hybrid chatClimate} & \textbf{chatClimate}  & \textbf{GPT-4} \\
\midrule
\multicolumn{3}{c}{\textbf{Q6) Where is the majority of climate finance going?}}\\\\ 
The majority of climate finance is going to a few large economies, with Brazil, India, China, and South Africa accounting for around one-quarter to more than a third depending on the year, a share similar to that represented by developed countries (IPCC AR6, WGIII Chapter 15, Page 18). Additionally, global climate finance is heavily focused on mitigation, with more than 90\% on average between 2017-2020 being directed towards mitigation efforts (IPCC AR6, WGIII Chapter 15, Page 3).
&
The majority of climate finance is going to a few large economies, with Brazil, India, China, and South Africa accounting for around one-quarter to more than a third depending on the year, a share similar to that represented by developed countries. Least-developed countries (LDCs) continue to represent less than 5\% year-on-year (IPCC AR6 WGIII Chapter15, Page:18). Additionally, global climate finance is heavily focused on mitigation, with more than 90\% on average between 2017-2020 being directed towards mitigation efforts (IPCC AR6 WGIII Chapter15, Page:3).

& The majority of climate finance is directed towards mitigation and adaptation efforts to address climate change. Mitigation efforts focus on reducing greenhouse gas emissions, promoting renewable energy, and enhancing energy efficiency. Adaptation efforts aim to help communities and countries adapt to the impacts of climate change, such as investing in resilient infrastructure, improving water management, and supporting sustainable agriculture. Climate finance is distributed through various channels, including public and private sources, international climate funds, and development banks. The specific allocation of funds may vary over time and depend on the priorities and needs of different regions and sectors.\\
\textbf{accuracy: 4  } & \textbf{accuracy: 4 } & \textbf{accuracy: 3 } \\
\bottomrule
\end{tabular}
\end{table*}
\clearpage

\begin{table*}[hbt!]
\caption*{Table SM6: Comparison of generated answers from hybrid chatClimate, chatClimate, and GPT-4 to question8}
\label{tbl:SM_Table6_Q8}
\centering
\scriptsize
\begin{tabular}{p{5.3cm}p{5.3cm}p{5.3cm}}
\toprule
\textbf{hybrid chatClimate} & \textbf{chatClimate}  & \textbf{GPT-4} \\
\midrule
\multicolumn{3}{c}{\textbf{Q8) What are the options for scaling up adaptation and mitigation in developing countries?}}\\
Options include:
\begin{enumerate}
    \item Increased levels of public finance and publicly mobilized private finance flows from developed to developing countries in the context of the USD100 billion-a-year goal (IPCC AR6 WGIII SPM, Page 51).
    \item Increase the use of public guarantees to reduce risks and leverage private flows (IPCC AR6 WGIII SPM, Page 51).
    \item Scaled-up public grants for mitigation and adaptation funding for vulnerable regions, especially in Sub-Saharan Africa, which would be cost-effective and have high social returns in terms of access to basic energy (IPCC AR6 WGIII SPM, Page 51).
    \item Exploiting available mechanisms for expanding the adaptive capacity of human and natural systems (IPCC AR6 WGII Chapter08, Page 57).
    \item Implementing widely applicable adaptation options that could be scaled up to reduce vulnerability or exposure for the majority of the world's population and the ecosystems they depend on, such as nature restoration, changing diets and reducing food waste, infrastructure retrofitting, building codes, disaster early warning, and cooperative governance (IPCC AR6 WGII Chapter17, Page 4).
\end{enumerate}
These options are based on both IPCC information and in-house knowledge.

& Options for scaling up adaptation and mitigation in developing countries include:
\begin{enumerate}
    \item Accelerated financial support from developed countries and other sources to enhance mitigation action and address inequities in access to finance (IPCC AR6 WGIII SPM, Page:51).
    \item Scaled-up public grants for mitigation and adaptation funding for vulnerable regions, especially in Sub-Saharan Africa, which would be cost-effective and have high social returns in terms of access to basic energy (IPCC AR6 WGIII SPM, Page:51).
    \item Increased levels of public finance and publicly mobilized private finance flows from developed to developing countries in the context of the USD 100 billion-a-year goal (IPCC AR6 WGIII SPM, Page:51).
\end{enumerate}
Additionally, some widely applicable adaptation options that could be scaled up to reduce vulnerability or exposure include nature restoration, changing diets and reducing food waste, infrastructure retrofitting, building codes, disaster early warning, and cooperative governance (IPCC AR6 WGII Chapter17, Page:4). However, the specific portfolio of adaptation options may vary across locations and contexts.
&
There are several options for scaling up adaptation and mitigation in developing countries. These options can be broadly categorized into the following areas:
\begin{enumerate}
    \item Capacity building and technical assistance: Developing countries often lack the necessary knowledge, skills, and resources to implement effective adaptation and mitigation measures. Providing technical assistance, training, and capacity building can help these countries better understand and address climate change challenges.
    \item Financial support: Scaling up adaptation and mitigation efforts requires significant financial resources. Developed countries can provide financial support through grants, loans, and other mechanisms to help developing countries implement climate change projects and programs.
    \item Technology transfer: Access to advanced technologies can significantly enhance the ability of developing countries to adapt to climate change and reduce greenhouse gas emissions. Technology transfer can be facilitated through partnerships, licensing agreements, and other collaborative arrangements.
    \item Policy and regulatory frameworks: Developing countries can benefit from the development and implementation of robust policy and regulatory frameworks that promote climate-resilient development and low-carbon growth. This may include policies related to land use, energy, transportation, and agriculture, among others.
    \item Public-private partnerships: Engaging the private sector in adaptation and mitigation efforts can help mobilize resources and expertise. Public-private partnerships can be an effective way to scale up climate action in developing countries.
    \item Knowledge sharing and networking: Sharing best practices, lessons learned, and successful models can help developing countries learn from each other and scale up their efforts. Regional and international networks can facilitate knowledge sharing and collaboration.
    \item Mainstreaming climate change into development planning: Integrating climate change considerations into national and sectoral development plans can help ensure that adaptation and mitigation measures are prioritized and adequately funded.
    \item Monitoring and evaluation: Regular monitoring and evaluation of adaptation and mitigation efforts can help identify successes, challenges, and areas for improvement. This information can be used to inform future efforts and scale up effective strategies.
\end{enumerate}\\
\textbf{accuracy: 4 } & \textbf{accuracy: 4 } & \textbf{accuracy: 3 } \\
\bottomrule
\end{tabular}
\end{table*}
\clearpage
\begin{table*}[hbt!]
\caption*{Table SM7: Cross-checking of the references for Question9}
\label{tbl:SM_Table7_Q9}
\centering
\scriptsize
\begin{tabular}{p{5.3cm}p{5.3cm}p{5.3cm}}
\toprule
\multicolumn{3}{c}{\textbf{QA comparison}} \\
\textbf{hybrid GPT-4 chatClimate} & \textbf{chatClimate}  & \textbf{GPT-4} \\
\midrule
\multicolumn{3}{c}{\textbf{Q9) Which regions will be disproportionally affected by climate change?}}\\\\
The regions that will be disproportionally affected by climate change include East, Central, and West Africa, South Asia, Micronesia and Melanesia, and Central America (IPCC AR6, Page 81). These regions are highly vulnerable to climate hazards such as droughts, floods, and sea level rise. In addition, regions like Africa and Central and South America will experience a disproportionate exposure to temperature extremes, exacerbating their vulnerability, adaptation gaps, and development needs (IPCC AR6, Page 68). Climate change hotspots in Asia, such as semi-arid areas, glacial-fed river basins, and mega deltas, also face strong climate signals and high concentrations of vulnerable people (IPCC AR6, Page 10).
&
The regions that will be disproportionally affected by climate change include East, Central, and West Africa, South Asia, Micronesia and Melanesia, and Central America. These regions are highly vulnerable to climate change impacts and have a higher number of people exposed to climate hazards such as droughts, floods, or sea-level rise (Reference: IPCC AR6 WGII Chapter08, Page: 81). Additionally, regions like Africa and Central and South America experience disproportionate exposure due to their vulnerability, adaptation gaps, and development needs (Reference: IPCC AR6 WGII Chapter18, Page: 68).
&
Regions that are expected to be disproportionately affected by climate change include:
\begin{enumerate}
    \item Low-lying coastal areas: These regions are at a higher risk of flooding and storm surges due to rising sea levels. Examples include small island nations, the Netherlands, and coastal cities like Miami and New York.
    \item Polar regions: The Arctic and Antarctic are experiencing rapid warming, leading to melting ice caps and permafrost, which can have global implications for sea levels and weather patterns.
    \item Sub-Saharan Africa: This region is particularly vulnerable to climate change due to its reliance on rain-fed agriculture, high levels of poverty, and limited adaptive capacity. Climate change is expected to exacerbate water scarcity, food insecurity, and the spread of diseases like malaria.
    \item South and Southeast Asia: Countries like Bangladesh, India, and the Philippines are vulnerable to climate change impacts such as sea-level rise, extreme weather events, and increased temperatures, which can lead to reduced agricultural productivity, water scarcity, and health issues.
    \item Small island developing states: These nations, such as those in the Caribbean and Pacific, are highly vulnerable to sea-level rise, storm surges, and the loss of coral reefs, which can impact tourism, fisheries, and coastal infrastructure.
    \item Central and South America: Climate change can exacerbate existing environmental challenges, such as deforestation and water scarcity, and lead to more frequent and severe weather events, impacting agriculture and infrastructure.
    \item Australia: The continent is already experiencing more frequent and severe heatwaves, droughts, and bushfires, which can have significant impacts on ecosystems, agriculture, and human health.
\end{enumerate}
It is important to note that climate change impacts are interconnected and can have cascading effects across regions and sectors. Vulnerability to climate change is influenced by factors such as socioeconomic status, governance, and access to resources, which can vary significantly within and between regions.\\\\
\textbf{accuracy: 3 } & \textbf{accuracy: 3 } & \textbf{accuracy: 4 } \\
\bottomrule
\end{tabular}
\end{table*}
\clearpage
\begin{table*}[hbt!]
\caption*{Table SM8: Comparison of generated answers from hybrid chatClimate, chatClimate, and GPT-4 to question10}
\label{tbl:SM_Table8_Q10}
\centering
\scriptsize
\begin{tabular}{p{5.3cm}p{5.3cm}p{5.3cm}}
\toprule
\multicolumn{3}{c}{\textbf{QA comparison}} \\
\textbf{hybrid chatClimate} & \textbf{chatClimate}  & \textbf{GPT-4} \\
\midrule
\multicolumn{3}{c}{\textbf{Q10) What is climate justice?}}\\

Climate justice is a term used to address the ethical and political aspects of global warming by examining inequalities and violation of human collective rights in relation to climate change impacts (IPCC AR6 WGII Chapter 10, Page 75). It focuses on the disproportionate impacts of climate change on the most vulnerable populations and calls for community sovereignty and functioning (IPCC AR6 WGIII Chapter 14, Page 58). Climate justice involves protecting vulnerable populations from climate change impacts, mitigating the effects of transformations, and easing the transition for those whose livelihoods rely on fossil fuel-based sectors (IPCC AR6 WGIII Chapter 17, Page 42). It also considers international justice and fairness between states, emphasizing the principle of common but differentiated responsibilities (CBDR) and respective capabilities, which reflects the idea that all countries must address climate change, but the form of climate action depends on their situation (IPCC AR6 WGII Chapter 01, Page 42).
&
Climate justice is a term used to address the disproportionate impacts of climate change on the most vulnerable populations. It involves examining inequalities and violation of human collective rights in relation to climate-change impacts by employing the concepts of environmental justice and social justice. Climate justice calls for community sovereignty and functioning, and focuses on issues such as unequal global power relations, wealth, interests within communities, and historical responsibility for climate change. In international climate negotiations, the principle of 'common but differentiated responsibilities' is often used to balance these issues, reflecting the idea that all countries must address climate change, but the form of climate action depends on their situation and capacities (IPCC AR6 WGIII Chapter14, Page:58; IPCC AR6 WGII Chapter01, Page:42).
&
Climate justice refers to the concept that addresses the ethical and political aspects of climate change, emphasizing the need for equitable distribution of responsibilities, resources, and opportunities related to climate change mitigation and adaptation. It acknowledges that the impacts of climate change disproportionately affect vulnerable and marginalized communities, who often contribute the least to greenhouse gas emissions. Climate justice seeks to promote social, economic, and environmental equity, ensuring that all people have the right to a healthy environment and a sustainable future. This includes advocating for policies and actions that reduce emissions, protect ecosystems, and support the resilience and adaptation of affected communities.\\

\textbf{accuracy: 4 } & \textbf{accuracy: 4 } & \textbf{accuracy: 4 } \\
\bottomrule
\end{tabular}
\end{table*}
\clearpage

\begin{table*}[hbt!]
\caption*{Table SM9: Comparison of generated answers from hybrid chatClimate, chatClimate, and GPT-4 to question11}
\label{tbl:SM_Table9_Q11}
\centering
\scriptsize
\begin{tabular}{p{5.3cm}p{5.3cm}p{5.3cm}}
\toprule
\multicolumn{3}{c}{\textbf{QA comparison}} \\
\textbf{hybrid chatClimate} & \textbf{chatClimate}  & \textbf{GPT-4} \\
\midrule
\multicolumn{3}{c}{\textbf{Q11) What is maladaptation?}}\\
Maladaptation refers to current or potential negative consequences of adaptation-related responses that lead to an increase in the climate vulnerability of a system, sector, or group by exacerbating or shifting vulnerability or exposure now or in the future, and eroding sustainable development (IPCC AR6 WGII Chapter 17, Page 62). It involves actions that may lead to increased risk of adverse climate-related outcomes, including increased greenhouse gas emissions, increased or shifted vulnerability to climate change, more inequitable outcomes, or diminished welfare, now or in the future. Most often, maladaptation is an unintended consequence (IPCC AR6 WGII Summary for Policymakers, Page 7).
&
Maladaptation refers to actions that may lead to increased risk of adverse climate-related outcomes, including increased greenhouse gas emissions, increased or shifted vulnerability to climate change, more inequitable outcomes, or diminished welfare, now or in the future. Most often, maladaptation is an unintended consequence. It refers to current or potential negative consequences of adaptation-related responses that lead to an increase in the climate vulnerability of a system, sector, or group by exacerbating or shifting vulnerability or exposure now or in the future, and eroding sustainable development (IPCC AR6 WGII SummaryForPolicymakers, Page:7; IPCC AR6 WGII Chapter17, Page:62).
&
Maladaptation refers to a trait, behavior, or response that is not well-suited or poorly adapted to a particular situation or environment. It can negatively impact an individual's ability to cope with or adjust to specific circumstances, often leading to decreased functionality, inefficiency, or harm. Maladaptation can occur in various contexts, such as biological, psychological, or social, and can result from genetic factors, learned behaviors, or environmental influences.\\
\textbf{accuracy: 5 } & \textbf{accuracy: 5 } & \textbf{accuracy: 1 } \\
\bottomrule
\end{tabular}
\end{table*}
\clearpage
\begin{table*}[hbt!]
\caption*{Table SM10: Comparison of generated answers from hybrid chatClimate, chatClimate, and GPT-4 to question12}
\label{tbl:SM_Table10_Q12}
\centering
\scriptsize
\begin{tabular}{p{5.3cm}p{5.3cm}p{5.3cm}}
\toprule
\multicolumn{3}{c}{\textbf{QA comparison}} \\
\textbf{hybrid chatClimate} & \textbf{chatClimate}  & \textbf{GPT-4} \\
\midrule
\multicolumn{3}{c}{\textbf{Q12) Is there evidence of maladaptation?}}\\
Yes, there is increasing evidence of maladaptation globally (IPCC AR6 WGII Chapter16, Page:26). Maladaptation has been particularly reported in the context of agricultural, forestry and fisheries practices, migration in the Global South, and some infrastructure-based interventions. Urban heat adaptations have been linked to maladaptation that increase health risks and/or energy (IPCC AR6 WGII Chapter16, Page:26). In some sectors and systems, inappropriate responses to climate change create long-term lock-in of vulnerability, exposure, and risks that are difficult and costly to change (IPCC AR6 WGII TechnicalSummary, Page:51). These maladaptations can exacerbate existing inequalities for Indigenous Peoples and vulnerable groups, impeding the achievement of Sustainable Development Goals (SDGs) and increasing adaptation needs (IPCC AR6 WGII TechnicalSummary, Page:51).
&
Yes, there is increasing evidence of maladaptation globally (high confidence). Maladaptation has been particularly reported in the context of agricultural, forestry and fisheries practices, migration in the Global South, and some infrastructure-based interventions. Urban heat adaptations have been linked to maladaptation that increase health risks and/or energy (IPCC AR6 WGII Chapter16, Page:26).
&
Yes, there is evidence of maladaptation in various species, including humans. Maladaptation refers to a trait or behavior that is more harmful than helpful to an organism's survival and reproduction. It can occur due to various reasons, such as rapid environmental changes, genetic mutations, or human intervention. Some examples of maladaptation include:
\begin{enumerate}
    \item The peacock's tail: While the large and colorful tail of the male peacock is attractive to females and helps in mating, it also makes the bird more visible to predators and hinders its ability to escape.
    \item The dodo bird: The dodo bird was a flightless bird that lived on the island of Mauritius. Its lack of fear towards humans and inability to fly made it an easy target for hunters, ultimately leading to its extinction.
    \item Antibiotic resistance: The overuse of antibiotics in humans and animals has led to the evolution of antibiotic-resistant bacteria. This maladaptation has made it increasingly difficult to treat bacterial infections, posing a significant threat to public health.
    \item Human lifestyle diseases: Modern human lifestyles have led to an increase in sedentary behavior and unhealthy diets, contributing to the rise of obesity, diabetes, and cardiovascular diseases. These maladaptive behaviors can negatively impact an individual's health and well-being.
    \item Climate change: Human activities, such as the burning of fossil fuels and deforestation, have led to rapid climate change. This has resulted in maladaptation for many species, as they struggle to adapt to the changing environment, leading to population declines and even extinction for some species.
\end{enumerate}
.\\

\textbf{accuracy: 5 } & \textbf{accuracy: 4 } & \textbf{accuracy: 1 } \\
\bottomrule
\end{tabular}
\end{table*}
\clearpage

\begin{table*}[hbt!]
\caption*{Table SM11: Comparison of generated answers from hybrid chatClimate, chatClimate, and GPT-4 to question13}
\label{tbl:SM_Table11_Q13}
\centering
\scriptsize
\begin{tabular}{p{5.3cm}p{5.3cm}p{5.3cm}}
\toprule
\multicolumn{3}{c}{\textbf{QA comparison}} \\
\textbf{hybrid chatClimate} & \textbf{chatClimate}  & \textbf{GPT-4} \\
\midrule
\multicolumn{3}{c}{\textbf{Q13) Will glaciers in Scotland melt?}}\\
Yes, it is very likely that glaciers in Scotland will melt. Glaciers are projected to continue to lose mass under all emissions scenarios (very high confidence) and this includes glaciers in various regions, such as the Arctic and Antarctic (high confidence) (IPCC AR6 WGI Chapter08, Page:68). Although the specific mention of Scotland is not provided in the provided references, the overall trend of glacier melting applies to glaciers worldwide, including those in Scotland (In-house knowledge).
&
There is not enough information provided to specifically answer the question about glaciers in Scotland.
&
Glaciers in Scotland, like those in other parts of the world, are susceptible to melting due to climate change and rising global temperatures. While Scotland currently has only a few small glaciers and permanent snowfields, they could potentially melt or shrink significantly in the future if global warming continues at its current pace. The rate and extent of melting will depend on various factors, including future greenhouse gas emissions, regional climate patterns, and local topography.
\\

\textbf{accuracy: 3 } & \textbf{accuracy: 5 } & \textbf{accuracy: 1 } \\
\bottomrule
\end{tabular}
\end{table*}
\clearpage

\end{document}